\documentclass[final]{l4dc2020} 
\usepackage{graphicx,caption,subcaption,xspace}
\usepackage{amsmath,amssymb,stmaryrd,mathtools}
\usepackage{algorithm}
\usepackage{algpseudocode}
\usepackage{acronym}
\usepackage{comment}
\usepackage{color}
\usepackage{units}
\usepackage{textcomp}

\usepackage[capitalise]{cleveref}

\DeclareCaptionLabelSeparator{periodspace}{.\quad}
\crefname{equation}{}{} % Default to eqref for equations
\crefname{section}{Sec.}{Sec.}

\newcommand{\R}{\mathbb{R}} % Real number
\newcommand{\state}{z} % state
\newcommand{\ctrl}{u} % control
\newcommand{\dstb}{d} % control
\newcommand{\pos}{p} % position

\newcommand{\cset}{\mathcal U}
\newcommand{\dset}{\mathcal D}
\newcommand{\cfset}{\mathbb U}
\newcommand{\dfset}{\mathbb D}

\newcommand{\ttr}{V_\text{R}} % time-to-reach
\newcommand{\ttc}{V_\text{C}} % time-to-collision
\newcommand{\goal}{\Gamma_G} % goal
\newcommand{\obst}{\Gamma_O} % obstacles

\renewcommand{\time}{t}
\newcommand{\image}{I} % state
\newcommand{\env}{\mathcal{E}}
\newcommand{\horizon}{H} % Horizon
\newcommand{\waypt}{\hat{w}} % waypoint

\renewcommand{\vec}[1]{\boldsymbol{#1}}

\newcommand{\metName}{LB-WayPtNav\xspace}

\algdef{SE}[SUBALG]{Indent}{EndIndent}{}{\algorithmicend\ }%
\algtext*{Indent}
\algtext*{EndIndent}

\newcommand{\dx}{d_{x}}
\newcommand{\dy}{d_{y}}
\newcommand{\dphi}{d_{\phi}}
\newcommand{\cost}{J}
\newcommand{\ttcscale}{\alpha}
\newcommand{\maxtime}{\bar{\ttc}}

\newcommand{\planhorizon}{N}
\newcommand{\timestep}{\Delta t}
\newcommand{\timepoint}{i}
\newcommand{\map}{M_{obs}}

\newcommand{\expertminobs}{d_{min}}

\newcommand{\expdataset}{S}

\newcommand{\ourcost}{ReachabilityCost\xspace}
\newcommand{\oldcost}{HeuristicsCost\xspace}
\newcommand{\ourcostNoDist}{ReachabilityCost-NoDstb\xspace}

\title[Generating Robust Supervision for Visual Navigation Using HJ Reachability]{Generating Robust Supervision for Learning-Based Visual Navigation Using Hamilton-Jacobi Reachability}
\usepackage{times}
\usepackage{wrapfig}

\usepackage{booktabs} 
\usepackage{textcomp}
\usepackage{diagbox}
\usepackage{colortbl}
\usepackage{amsmath}
\usepackage{amssymb}
\usepackage{bm}

\author{%
 \Name{Anjian Li}\textsuperscript{1} \Email{anjianl@sfu.ca}
 \AND
 \Name{Somil Bansal}\textsuperscript{2} \Email{somil@berkeley.edu}
 \AND
 \Name{Georgios Giovanis}\textsuperscript{1} \Email{ggiovani@sfu.ca}
 \AND
 \Name{Varun Tolani}\textsuperscript{2} \Email{vtolani@berkeley.edu}
 \AND
 \Name{Claire Tomlin}\textsuperscript{2} \Email{tomlin@berkeley.edu}
 \AND
 \Name{Mo Chen}\textsuperscript{1} \Email{mochen@sfu.ca} \\
 \addr  \textsuperscript{1}School of Computing Science,\quad\quad\quad \textsuperscript{2}Department of Electrical Engineering and Computer Sciences,\\
 Simon Fraser University,\quad\quad\quad  ~~~~~~~~~~~University of California, Berkeley,\\
 Burnaby BC, Canada, V5A 1S6\quad\quad\quad
 Berkeley, CA 94720, USA
}

\makeatletter
 \let\Ginclude@graphics\@org@Ginclude@graphics 
\makeatother

\begin{document}

\maketitle

% Abstract
% \begin{abstract}%
% Visual navigation for autonomous robots is appealing due to the availability of low-cost cameras.
% End-to-end learning has been a popular approach but requires a large amount of data and lacks the generalizability to the real world.
% In contrast, optimal control does not require data and is generalizable, but requires full state information and \textit{a prior} known map.
% To take advantages from both classes of methods, we adopt a hybrid approach in which a convolutional neural network (CNN) predicts waypoints and optimal control is used to generate trajectories to the waypoints; this framework enables successful simulation-to-real world deployment.
% In this paper, we present a novel method that uses reachability analysis to generate training data for waypoint prediction. 
% Our method implicitly encodes system dynamics information in the CNN to predict efficient and collision-free waypoints.
% Through incorporating disturbances in dynamics, our method is robust to CNN prediction errors.
% We demonstrate superior performance on navigating through narrow spaces in both simulation and the real world. 
% \end{abstract}

\begin{abstract}%
In \cite{bansal2019combining}, a novel visual navigation framework that combines learning-based and model-based approaches has been proposed. 
Specifically, a Convolutional Neural Network (CNN) predicts a waypoint that is used by the dynamics model for planning and tracking a trajectory to the waypoint. 
However, the CNN inevitably makes prediction errors which often lead to collisions in cluttered and tight spaces. 
In this paper, we present a novel Hamilton-Jacobi (HJ) reachability-based method to generate supervision for the CNN for waypoint prediction in an unseen environment. 
By modeling CNN prediction error as ``disturbances'' in robot's dynamics, our generated waypoints are robust to these disturbances, and consequently to the prediction errors. Moreover, using globally optimal HJ reachability analysis leads to predicting waypoints that are time-efficient and avoid greedy behavior. Through simulations and hardware experiments, we demonstrate the advantages of the proposed approach on navigating through cluttered, narrow indoor environments.
\end{abstract}


% Introduction
\section{Introduction}

Autonomous navigation is fundamental to control and robotics.
Following the success of deep learning, visual navigation has gained popularity.
One appeal of visual navigation -- which involves using one or more cameras and computer vision to perceive the environment to reach navigational goals -- is that cameras are cheap, light weight, and ubiquitous. 
%This makes the deployment of a large number of autonomous robots more scalable.

Typically, a geometric map of the environment is used for navigation~(\cite{thrun2005probabilistic,slam-survey:2015, lavalle2006planning}). However, real-time map generation can be challenging in texture-less environments or in the presence of transparent, shiny objects, or strong ambient lighting~(\cite{alhwarin2014ir}). 
In contrast, end-to-end (E2E) learning approaches have been used for locomotion~(\cite{gandhi2017learning,kahn2017uncertainty,sadeghi2016cadrl,kang2019generalization}) and goal-point navigation~(\cite{zhu2016target, gupta2017cognitive, khan2017memory, kim2015deep, pan2017agile}) that side-step this explicit map estimation step, but suffer from data inefficiency and lack of robustness~(\cite{recht2018tour}). Consequently, a number of papers seek to combine the best of learning with optimal control for high-speed navigation~(\cite{richter2018bayesian, jung2018perception, loquercio2018dronet, bansal2018chauffeurnet, muller2018driving, meng2019neural}), race-track driving~(\cite{pmlr-v78-drews17a, drews2019vision}), and drone racing~(\cite{kaufmann2018beauty, kaufmann2018deep}). In particular, \cite{bansal2019combining} combines the ideas from optimal control and computer vision by having a convolutional neural network (CNN) predict waypoints instead of control signals and using optimal control to obtain the control for reaching the waypoints. 
This hybrid approach greatly improves generalizability: a CNN trained in simulation could be successfully deployed on a real robot without any additional training or tuning. 
However, the inevitable errors in waypoint predictions during the test time may cause unintended robot trajectories, ultimately resulting in collisions with the obstacles. 
This is particularly problematic when the robot needs to navigate through cluttered environments or narrow openings, as the error margin in such scenarios is often small.

% End-to-end (E2E) visual navigation has been studied in scenarios such as autonomous driving (\cite{?}), and involves learning a direct mapping from images to control signal.
% Due to poor data efficiency, E2E methods are only effective when a large amount of data is available.
% %Obtaining such data is often challenging for physical systems.
% %In addition, control policies trained from E2E methods often over-specialize to specific tasks, and robustness of the learned policy is often inadequate.
% In addition, the lack of generalizability and robustness of learned policies makes training in simulation and testing in the real world challenging.

%%%%%%%%%%%%%%%%%%%%%%%%%%%%%%%%%%%%%%%%%%%%%%%%%%%%%%%%%%%%%%%%%%%%%%%%%%%%%%%%%%%%%%%%%%%%%%%%%%%
\begin{wrapfigure}{r}{0.4\textwidth}
% \vspace{4pt}
\vspace{-2em}
  \begin{center}
    \includegraphics[width=\linewidth]{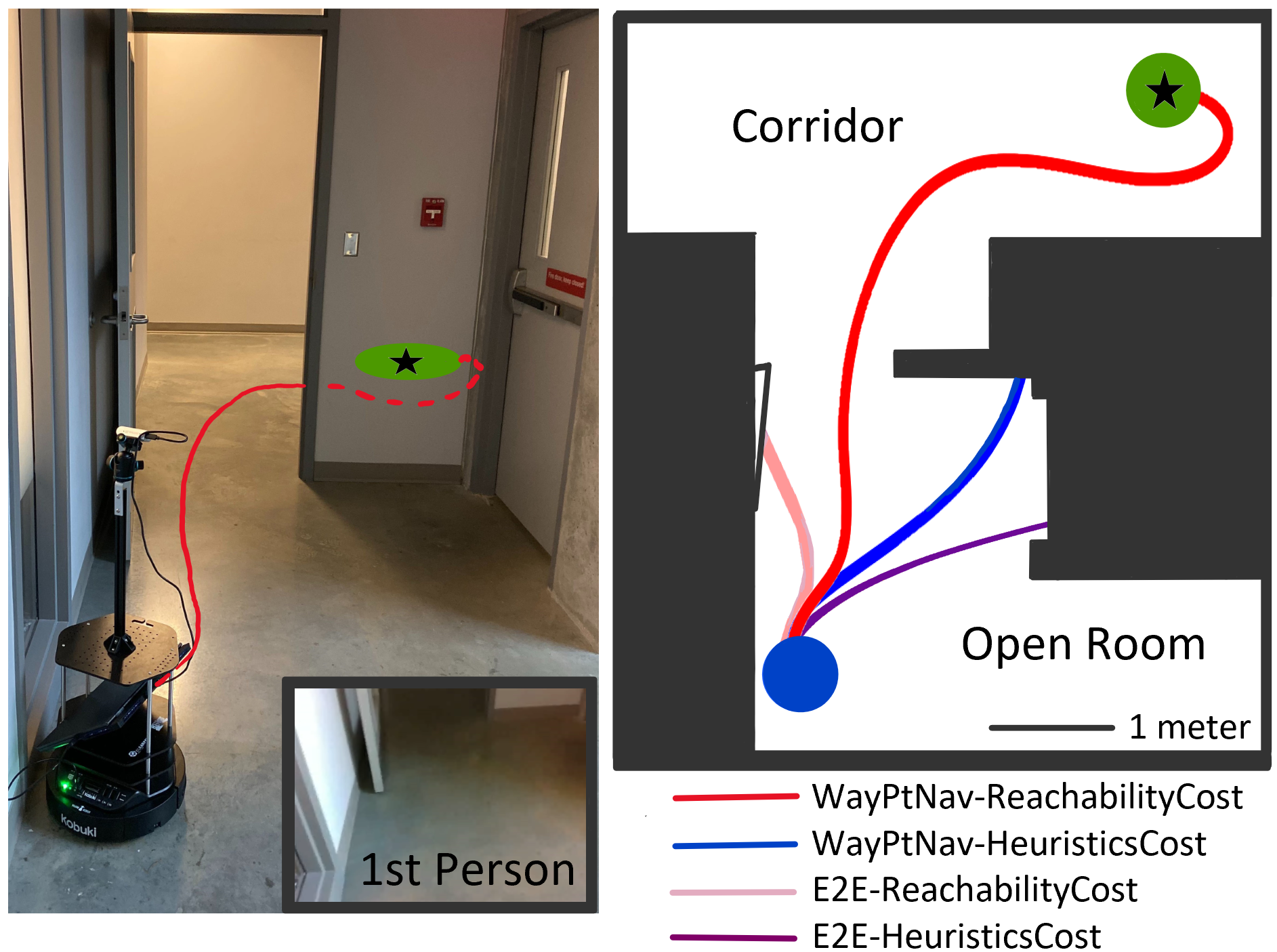}
  \end{center}
  \vspace{-1.5em}
  \caption{Hardware experiment 1. On the left, 3rd and 1st person views of the Turtlebot2 testbed are shown along with a trajectory (red) to the goal (green). On the right, a birds-eye view of the environment is displayed. The Turtlebot's starting location is shown as a blue circle. }
  \label{fig:Scenario_1}
  \vspace{-0.5em}
% \vspace{4pt}
\end{wrapfigure}
%%%%%%%%%%%%%%%%%%%%%%%%%%%%%%%%%%%%%%%%%%%%%%%%%%%%%%%%%%%%%%%%%%%%%%%%%%%%%%%%%%%%%%%%%%%%%%%%%%%

\noindent \textbf{Contributions}: 
Based on the framework in \cite{bansal2019combining} we propose a novel reachability-based method to generate robust waypoints for supervising the CNN.
Our key insight is to model the CNN prediction error as ``disturbance'' in robot's dynamics and generate waypoints that are optimal under worst-case disturbances, ensuring robustness despite the prediction errors.
Compared to other works in the safe learning literature such as \cite{Fisac2019} and \cite{Bajcsy2019}, which wrap reachability-based safety controllers around policies being learned, we provide an alternative that pre-emptively uses disturbances in training data generation.

Unlike \cite{bansal2019combining}, which relies on distance-based heuristics, our method involves solving static Hamilton-Jacobi (HJ) (\cite{Bansal2017a}) partial differential equations (PDEs).
The obtained value functions represent the time until goal-reaching and time until collision despite worst-case disturbances, given system dynamics and a known environment.
These value functions are combined into a cost map that precisely quantifies the quality of waypoints and considers all possible combinations of states by construction.
This leads to less greedy navigation behavior and significant improvement in the success rate during test time.

% Crucially, our method leverages reachability analysis to account for worst-case disturbances.
%Disturbances traditionally model environmental factors such as wind and other agents.
% In this paper, one key insight is that the use of disturbances can make training CNNs more robust.

Overall, our approach leads to less greedy navigation behaviors that are robust to CNN prediction errors.
Through simulations and real-world experiments, we demonstrate that practical scenarios such as safely moving through narrow doorways become possible with the proposed approach.

\section{Problem Setup}
We consider the problem of autonomous navigation in \textit{a priori} unknown static environment. %, under the assumption of perfect odometry.
Starting from an initial position, the robot needs to reach a goal position $\pos^* = (x^*, y^*)$. 
We model our ground vehicle as a four-dimensional (4D) system with the following dynamics:
\begin{equation} \label{eqn:NumSimpleDyn}
\dot{x} = v\cos\phi,\quad \dot{y} = v\sin\phi,\quad \dot{v} = a,\quad \dot{\phi} = \omega\,,
\end{equation}
\noindent where the state $\state(\time)$ at time $\time$ consists of the position $(x(\time), y(\time))$, speed $v(\time)$, and heading $\phi(\time)$.
The control is acceleration and turn rate, $\ctrl(\time) := (a(\time), \omega(\time))$.
The robot is equipped with a forward-facing, monocular RGB camera mounted at a fixed height and oriented at a fixed pitch.
At time $\time$, the robot receives an RGB image of the environment $\env$, $\image(\time) = \image(\env, \state(\time))$, the state $\state(\time)$, and the target position $\pos^* = (x^*, y^*)$.
The objective is to obtain a control policy that uses these inputs to guide the robot to within a certain distance of $\pos^*$.

%\mcnote{Is there a reason for the subscript $t$? $\state(t)$ would be more consistent with typical notation in continuous-time control.}

% Background
\section{Background}
We build upon the learning-based waypoint approach to navigation (\metName{}) proposed in \cite{bansal2019combining}. 
However, unlike \metName{}, we use a HJ Reachability-based framework to generate supervision data.
We now provide a brief overview of \metName{} and HJ reachability.

\subsection{\metName{}}
\begin{wrapfigure}{r}{0.45\textwidth}
\vspace{-4em}
  \begin{center}
    \includegraphics[width=\linewidth]{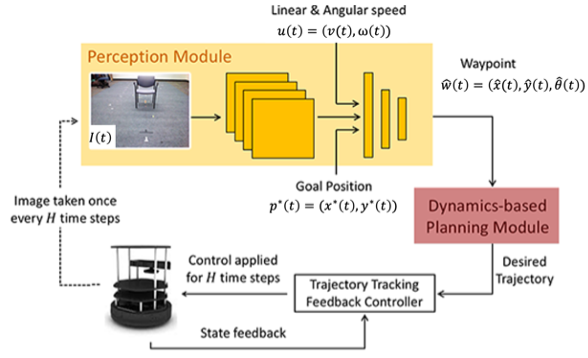}
  \end{center}
  \vspace{-2em}
  \caption{\metName{} framework adapted from \cite{bansal2019combining}. %\SBnote{This diagram needs to be modified slightly as $u_t$ definition has changed.} %\mcnote{This is the background section, so strictly speaking we should stick to the old material. Perhaps just have a generic $u_t$?}
  }
  \label{fig:lbwayptnav_framework}
  \vspace{-1.5em}
\end{wrapfigure}
\metName{} combines a learning-based perception module with a dynamics model-based planning and control module for navigation in \textit{a priori} unknown environments (see Fig. \ref{fig:lbwayptnav_framework}). 
\smallskip
\\
\textbf{Perception module.} The perception module is implemented as a CNN that takes as input a $224 \times 224$ pixel RGB image, $\image(\time)$, captured from the onboard camera, the target position, $\pos^*$, specified in the vehicle's current coordinate frame, and vehicle's current linear and angular speed, $(v(\time), \omega(\time))$, and outputs the desired state or a waypoint $\waypt(\time) := (\hat{x}(\time), \hat{y}(\time), \hat{\theta}(\time))$.
\smallskip
\\
\textbf{Planning and control module.} Given a waypoint $\waypt(\time)$, the system dynamics in Eqn.~\eqref{eqn:NumSimpleDyn} are used to plan a spline-based trajectory to $\waypt(\time)$, starting from the current state of the vehicle.
This leads to a smooth, dynamically feasible, and computationally efficient trajectory to the waypoint.
An LQR-based feedback controller tracks the trajectory. %, $\{k([\time,\time+\horizon]),~K([\time,\time+\horizon])\}~=~LQR(z^*([\time,\time+\horizon]),~u^*([\time,\time+\horizon]))$.
%Here $k$ and $K$ represent the feed-forward and feedback terms respectively.
%The LQR controller is obtained using the dynamics in Eqn.~\eqref{eqn:NumSimpleDyn}, linearized around the trajectory $\{z^*, u^*\}$. 
%\smallskip
The commands generated by the LQR controller are executed on the system for $\horizon$ seconds, and then a new image is used to generate another waypoint and trajectory. 
%\mcnote{This part sounds like the replanning frequency is the same as the planning horizon}
%Consequently, a new waypoint and plan are generated.
This process is repeated until the robot is within a certain distance of $\pos^*$.

\subsection{The Time-to-Reach Problem in Reachability Analysis}
Consider a dynamical system described by
$\dot{\state}\left(t\right) =f\left(\state(\time),\ctrl(\time), \dstb\left(t\right)\right), \state\left(0\right)  =\state_0$,
where $\state\in\R^{n}$ is the state, and $u\in\cset, d\in\dset$ are the control and disturbance respectively.
In this paper, the control represents the actions a robot can take to change its state, and the disturbance primarily models CNN prediction errors.
%We assume $f:\R^{n}\times\cset\times\dset\rightarrow\R^n$ is uniformly continuous, bounded, and Lipschitz continuous in $\state$ for fixed $\ctrl$ and $\dstb$, and $\ctrl\left(\cdot\right)\in\cfset$, $\dstb\left(\cdot\right)\in\dfset$ are measurable functions, so that \eqref{eq:syst-gen} has a unique solution.
%We also assume that the disturbance $\dstb(\cdot)\in\dfset$ is restricted to using only \textit{non-anticipative} strategies as in \cite{Mitchell05}.
%\ajnote{We are using reach-avoid ttr. Is this correct then?}
%In the time-to-reach (TTR) problem, one computes the time it takes to reach a target set $\Gamma$ from any initial state $\state$, possibly while satisfying state constraints. 
Following \cite{Yang2013} and \cite{Takei10}, we define the time-to-reach (TTR) value function denoted as $\ttr(\state)$, which represents the minimum time required to drive the system to the goal set $\goal$ while avoiding the obstacle set $\obst$, despite worst-case disturbance.
The control minimizes this time and disturbance maximizes: $ \ttr\left(\state\right)=\max_{\dstb(\cdot)\in\dfset} \min_{\ctrl(\cdot)\in\cfset} \min\{ t| \state(\time)\in\goal\wedge \forall s \in [0, t], \state(s) \notin \obst\}$.
We also define the time-to-collision (TTC) value function, $\ttc(\state)$, which represents the maximum time until collision with the obstacle set $\obst$ assuming that the control is optimally avoiding obstacles under worst-case disturbance: $\ttc(\state)=\min_{\dstb(\cdot)\in\dfset}\max_{\ctrl\in\cfset}\min\{ t| \state(\time)\in\obst\}$.
Applying the dynamic programming principle, we can obtain $\ttr(\cdot)$ and $\ttc(\state)$ respectively as the viscosity solution to the following stationary HJ PDEs:
\begin{align}
&\ttr(\state) =0\text{ in }\goal, \quad
\ttr(\state)  = \infty \text{ in }\obst, \quad 
\max_{\ctrl\in\cset} \min_{\dstb\in\dset}\{ -\nabla\ttr(\state)^\top f(\state,\ctrl,\dstb)-1\}  =0 \text{ otherwise} \nonumber \\
&\ttc\left(\state\right) =0 \text{ in }\obst, \quad
\min_{\ctrl\in\cset} \max_{\dstb\in\dset}\left\{ -\nabla\ttc\left(\state\right)^\top f\left(\state,\ctrl,\dstb\right)-1\right\}  =0 \text{ otherwise} \label{eq:HJPDEs}
\end{align}

% Reachability-based supervision
\section{Reachability-based Supervision for Waypoints} \label{sec:reachability_base_supervision}
%\mcnote{General framework goes here, and elaborated in subsections; make sure to reuse notation from earlier sections}

% \ajnote{1210:\\
% Talk about how to collect data for imitation learning: images are rendered, and waypoints $z_{I+H}$ are supervised by reachability expert.\\
% Introduce some notation here\\
% Have a diagram to show the work flow of our method (1. Compute TTR, TTC, 2. Solve MPC, get waypoint pairs 3. Train NN)
% }

%\ajnote{Use $\timepoint$ for discrete time point, use $\time$ for continuous time}

% %%%%%%%%%%%%%%%
% \begin{wrapfigure}{r}{0.45\textwidth}
% \vspace{-2em}
% \begin{center}
% \includegraphics[width=1\linewidth]{figures/flow-chart_v3.png}
%     \end{center}
%     \vspace{-2em}
%     \caption{Workflow of training data generation using reachability expert.}
%     \label{{fig:flow_chart}}
%     \vspace{-1em}
% \end{wrapfigure}
% %%%%%%%%%%%%%%%

In the perception module presented in Fig. \ref{fig:lbwayptnav_framework}, from the start position, the robot sequentially observes the image $\image(\time)$, linear speed $v(\time)$ and angular speed $\omega(\time)$ at time $\time$ to predict a waypoint $\waypt$ with a CNN.
This waypoint prediction is conducted at a fixed replan frequency until the robot reaches the goal $\goal$, defined to be positions within certain distance to $\pos^*$.

\begin{figure}
    \vspace{5pt} 
    \centering
    \includegraphics[width=0.7\linewidth]{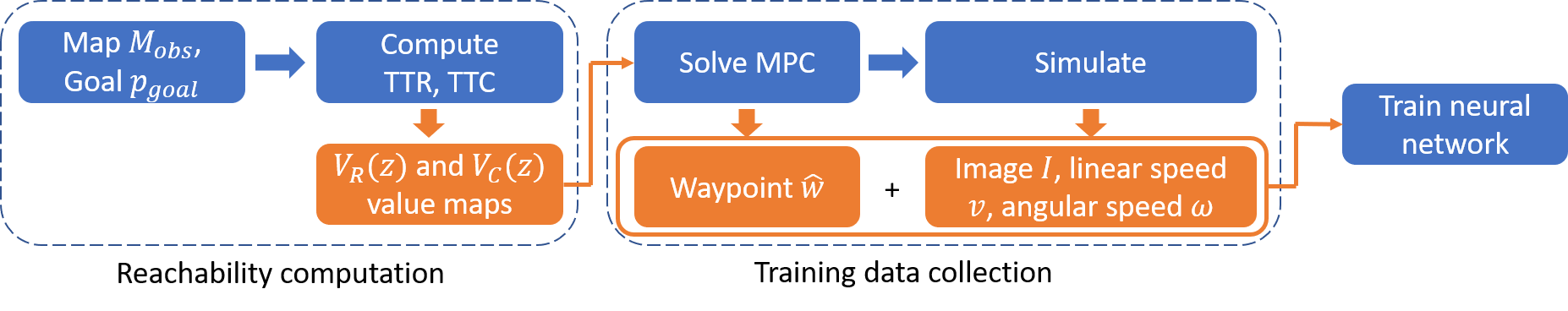}
    \vspace{-1em}
    \caption{Workflow of training data generation using reachability expert.}
    \label{fig:flow_chart}
\vspace{-1em}
\end{figure}

To generate supervision for safe and efficient waypoints, we propose a novel reachability expert to autonomously navigate in simulation and collect training data.
% For the perception module presented in Fig. \ref{fig:lbwayptnav_framework}, we propose a novel reachability expert to automatically accomplish the navigation tasks in simulation and generate supervision for waypoints.
% For the perception module, we present a reachability-based imitation learning scheme to generate supervision for waypoints.
% Instead of having human experts, we design a reachability expert to automatically accomplish the navigation tasks in simulation and collect waypoints $\waypt(\time)$, current image $\image(\time)$, and linear and angular speed $(v(\time), \omega(\time))$ as training data.
Specifically, given an obstacle map $\map$ and a goal area $\goal$, one can compute corresponding TTR and TTC value maps, which are integrated in the cost function of a model predictive control (MPC) optimization problem.
% Then, the reachability expert adopts a model predictive control (MPC) framework that integrates TTR and TTC values as the cost function.
By solving this MPC problem, the optimal waypoint $\waypt$ is obtained, and at time $\time$, the image $\image(\time)$ is rendered, and linear and angular speed $\{v(\time), \omega(\time)\}$ are measured in simulation environment.
% This data-label pair $\{(\image, v, \omega),  \waypt\}$ is used to train the CNN in Fig. \ref{fig:lbwayptnav_framework}.
Finally, we repeat the above procedure in different navigation tasks until sufficient data-label pairs $\{(\image, v, \omega),  \waypt\}$ are obtained for the training dataset.
The entire procedure is illustrated in Fig. \ref{fig:flow_chart}.

After training, the neural network can robustly transfer to novel, unknown environments.

\subsection{TTR and TTC Computations}

%\mcnote{This is new to our method}
% \ajnote{Are we using $x(t)$ or $x$}

% \ajnote{1210:\\
% TTC bar not Tmax, show something related to TTC
% }

We add disturbances to Eqn. \eqref{eqn:NumSimpleDyn} to describe the expert's system dynamics
\begin{align} 
&\dot{x} = v\cos\phi + \dx,\quad \dot{y} = v\sin\phi + \dy, 
\quad \dot{v} = a,\quad \dot{\phi} = \omega\ + \dphi, \label{eqn:ExpSysDyn} \\
&v \in [0, \bar v], \quad a \in [-\bar a,\bar a ], \quad \omega \in [- \bar \omega, \bar \omega], 
 \quad d_{x}^{2} + d_{y}^{2} \leq \bar d_{xy}^{2}, \quad \dphi \in [- \bar \dphi, \bar \dphi] \label{eq:constraints}
\end{align}

\noindent with states $\state = (x, y, v, \phi)$, controls $\ctrl = (a, \omega)$ and disturbances
% are added on the derivatives of positions and orientation, 
$\dstb = (\dx, \dy, \dphi)$.
Also, $\bar a$, $\bar \omega$, and $\bar d_{\phi}$ are upper bounds for acceleration, angular speed and disturbances in turn rate, and $\bar d_{xy}$ is the circular upper bound for disturbances in $x$ and $y$ components of speed.

For every navigation task, we initialize a goal position $\pos^*$ on an known obstacle map $\map$. 
We define the goal area $\goal$ to be $\goal=\{(x,y,\phi,v): \sqrt{(x-\pos^*)^2+(y-\pos^*)^2}\le c\}$, and the map $\map$ as $\obst$.
$\ttr({\state})$ and $\ttc({\state})$ are then computed based on Eqn. \eqref{eq:HJPDEs}.
$\ttr({\state})$ and $\ttc({\state})$ guide the reachability expert in a 4D state space for goal reaching and collision avoidance. %Please refer to \cite{Yang2013} for more computation details.

%As described in Eqn.\eqref{eq:HJPDE} and Eqn.\eqref{eq:HJPDEobs}, 
Incorporating worst-case disturbances leads to more conservative values of $\ttr({\state})$ and $\ttc({\state})$, which result in less greedy expert trajectories in a nuanced manner. Greediness often makes the robot incapable of going around obstacles and having a view of the environment that informs the robot about possible routes to the goal. %\vtnote{The above few sentences feel a bit vague- I know what you are trying to say, but somebody who is unfamiliar with this project may not know what you mean by "a good view"- if you have space a picture of what is a good view and what isn't might be helpful here.}
Crucially, the conservatively safe trajectories address the prediction errors from neural networks, since a minor deviation will not lead to collision.
Note that $\map$ is assumed to be known only during training; no such assumption is made during the test time when the robot only relies on onboard sensors for navigation.

\subsection{Waypoint Supervision Generation} \label{sec:waypt_generation}
%\mcnote{This is new to our method}

% \ajnote{1210: Combine 4.2 and 4.3, as a section of waypoint navigation}

We use an MPC framework to generate waypoints and expert trajectories.
To achieve efficient and safe navigation, we trade off between reaching the goal faster and staying further from obstacles. 
Thus we design a novel cost function, \ourcost $\cost$, to be a combination of TTR and TTC:
\begin{equation} \label{eq:CostFun}
    \cost(\state) = \ttr(\state) + \ttcscale(\maxtime -\ttc(\state)),
\end{equation}

\noindent where $\maxtime$ is the computational upper bound for TTC, and $\ttcscale$ is the scale factor. 

The expert uses MPC to plan, in a receding horizon manner, a trajectory of time horizon $\horizon$ until the goal is reached. 
%and replans in a shorter horizon $\replanhorizon < \horizon$, until reaching the goal area.
Starting from $t=0$, in every MPC optimization problem, we discretize the time horizon $[\time, \time+\horizon]$ to be $\{\time_i | \time + i\Delta t, i\in\{0,1,...,\planhorizon\} \}$. %, where $\horizon$ is determined by replan frequency. 
At any time index $\timepoint$ 
% \in \{0,1,...,\planhorizon\}$
, we denote $\state^{[\timepoint]}:=\state(\time_i)$ and $\cost^{[\timepoint]}:=\cost(\time_i)$.
% Then reachability expert sequentially solves the following MPC problem in $[\time, \time+\horizon]$:
Then the following MPC problem is sequentially solved in $[\time, \time+\horizon]$:
\begin{align} 
    \underset{\tiny
    \begin{aligned} \label{eq:MPC}
    \state^{[0]},...,\state^{[N]}, \nonumber \\ \ctrl^{[0]},...,\ctrl^{[N]}
    \end{aligned}}
    {\text{minimize }}& %\min_{\mathbf{\state}, \mathbf{\ctrl}}
    %\cost(\mathbf{\state}), \quad \cost(\mathbf{\state}) = 
    \sum_{\timepoint=0}^{\planhorizon} \cost^{[\timepoint]}(\state^{[\timepoint]})
    \\
    \text{subject to } 
    & x^{[\timepoint+1]} = x^{[\timepoint]} + \timestep v^{[\timepoint]} \cos{\phi^{[\timepoint]}}, \quad y^{[\timepoint+1]} = y^{[\timepoint]} + \timestep v^{[\timepoint]} \sin{\phi^{[\timepoint]}}, \quad  \phi^{[\timepoint+1]} = \phi^{[\timepoint]} + \timestep w^{[\timepoint]}, \nonumber \\ 
    & v^{[\timepoint+1]} = v^{[\timepoint]} + \timestep a^{[\timepoint]},
    \quad v^{[\timepoint]} \in [0,\bar v], \quad \omega^{[\timepoint]} \in [- \bar \omega, \bar \omega], \nonumber \\ 
    % & z^{[0]} = (x(\time),y(\time),v(\time),\phi(\time)), \quad z^{[N]} = (x(\time+\horizon),y(\time+\horizon),v(\time+\horizon),\phi(\time+\horizon)), \nonumber \\
    % & \ctrl^{[0]} = (a(\time),\omega(\time)), \quad \ctrl^{[N]} = (a(\time+\horizon),\omega(\time+\horizon)) \nonumber
    % For simplification, we can use the following
    & z^{[0]} = z(\time), \quad z^{[N]} = z(\time + \horizon),
    \quad \ctrl^{[0]} = \ctrl(\time), \quad \ctrl^{[N]} = \ctrl(\time + \horizon)
\end{align}
%where $\bm{\state} := (\state^{[0]},...,\state^{[N]})$ and $\mathbf{\ctrl}:= (\ctrl^{[0]},...,\ctrl^{[N]})$.

To solve Eqn. \eqref{eq:MPC} in $[\time,\time+\horizon]$, the reachability expert first samples a local waypoint grid $\bm{\waypt}$ in the heading direction
% robot's own coordinate frame within current field of view.
% The expert considers the grid 
as possible final states $z^{[N]}$: $\bm{\waypt} := (\bm{\hat x}^{[N]}, \bm{\hat y}^{[N]}, \bm{\hat \phi}^{[N]})$, and compute dynamically feasible spline trajectories to each waypoint using differential flatness of Eqn. \eqref{eqn:ExpSysDyn} (\cite{WALAMBE2016601}). 
%Due to the constraints of initial and final states, the trajectory is unique for each waypoint.
Next, the expert filters out the invalid waypoints whose trajectory violates control constraints.
% Finally, the expert chooses the optimal waypoint $\waypt^{[\timepoint]}$ with the solution trajectory $\bm{\state}$ to be the spline trajectory that has the minimal cost of $\cost^{[\timepoint]}$ for .
Finally, the solution trajectory $\bm{\state}$ with the minimum cost is chosen, and the corresponding waypoint $\waypt=(\hat x^{[N]}, \hat y^{[N]}, \hat \phi^{[N]})$ is added to the training data set along with the image $\image(t)$ and speeds $v(t), \omega(t)$ at time $\time$.
By solving the MPC problem many times, we obtain the expert dataset $\expdataset = \{(I_k(t), v_k(t), \omega_k(t)), (\hat x^{[N]}_k, \hat y^{[N]}_k, \hat \phi^{[N]}_k)\}_{k=1}^{M}$, where $k$ is the index of the MPC problem, and $M$ is the total number of data points (and the total number of MPC problems solved).

\section{Summary of Simulation Results} \label{sec:simulation_results}
After training on the generated expert dataset $\expdataset$, we test our model in a novel environment in simulation without an \textit{a priori} known map.

% $x^{[i]}_{k}$

% $\{(I^{[0]}_k, v^{[0]}_k, \omega^{[0]}_k), (x^{[H]}_k, y^{[H]}_k, \theta^{[H]}_k)\}_{k=1}^{M}$

\noindent \textit{\textbf{Dataset:}} We use Stanford large-scale 3D Indoor Spaces Dataset \cite{armeni20163d} as our simulation environment, which are 3D scans of real world buildings.
Two buildings are used for data generation and training; the 3rd \textit{held-out} building is used as the test environment, which has significant differences in the object appearance and layout. 
For navigation tasks in training and testing, we sample various start and goal positions that require the robot to go through narrow openings.

\noindent \textit{\textbf{Implementation details:}} We train the CNN in Fig. \ref{fig:lbwayptnav_framework} with 150k data points from the reachability expert, $M=150$k. 
The mean squared error loss is used and optimized using the Adaptive Moment Estimation (ADAM) algorithm with a learning rate of $10^{-4}$ and weight decay of $10^{-6}$.

\noindent \textit{\textbf{Metrics:}} We use both statistics and trajectory plots to present the test results.
For statistics, we use success rate, average time to reach the goal area (for successful tasks), acceleration and jerk to measure the quality of trajectories. 
With trajectory plots, we analyze the robot's specific behaviors.

\noindent \textit{\textbf{Baselines:}} 
We compare our approach with the \oldcost designed in \cite{bansal2019combining} for the MPC framework:
\begin{equation} \label{eq:oldcost}
    J^{heuristic}(\state) := (\max \{0, \lambda_{1} - d^{obs}(x,y) \})^{3} + \lambda_{2}(d^{goal}(x,y))^{2}
\end{equation}

\noindent where $d^{obs}(x,y)$ is the distance to the nearest obstacle, $d^{goal}(x,y)$ is the distance to the goal, and $\lambda_{1}$ and $\lambda_{2}$ are scaling factors.
We compare our \ourcost in Eqn. \eqref{eq:CostFun} to the \oldcost in Eqn. \eqref{eq:oldcost} for two different navigation frameworks: waypoint navigation (WayPtNav) and end-to-end (E2E) learning, resulting in a total of 4 methods.
WayPtNav maps the current image and controls inputs to the waypoint (as in Fig. \ref{fig:lbwayptnav_framework}), where E2E learning directly outputs the control commands given the same inputs.
We also compare against an additional baseline, \ourcostNoDist, which does not incorporate disturbances in the system dynamics during the data generation.
% 
% \vtnote{While this has all the information needed I found the wording a bit cumbersome- I was actually unclear what the baselines were until looking at table 1. If you have space it might be helpful to add one more sentence to the effect of "in total we report metrics for 5 methods WayPtNav-ReachabilityCost, WayPtNav-HeuristicsCost, ...}
% \vtnote{Also, is there a reason there is no number for E2E-ReachabilityCost without disturbances? Readers might wonder why this is not here for E2E since it is for WayPtNav.}

% Expert performance
\subsection{Expert Performance} \label{sec:expert_performance}
%%%%%%%%%%%%%%%%%%%%%%%%%%%%%%%%%%%%%%%%%%%%%%%%%%%%%%%%%%%%%%%%%%%%%%%%%%%%%%%%%%%%%%%%%%%%%%%%%%%
% \begin{wrapfigure}{r}{0.5\textwidth}
% % \vspace{5pt}
% \vspace{-4em}
%   \begin{center}
%     \includegraphics[width=\linewidth]{figures/old_ours_expert_traj_v3.png}
%   \end{center}
%   \vspace{-2em}
%   \caption{Expert trajectories obtained using. 
%   (a) Baseline expert; (b) Reachability expert. 
%   Baseline expert starts with a greedier path close to the wall, and fails to enter the narrow opening due to the hard obstacle padding (medium grey) used during waypoint optimization. 
%   Reachability expert chooses better combination of positions, orientation and speed as waypoints that are conservatively away from obstacles and safely go through the opening.}
% %   \vtnote{It might be informative to actually show the hard margins on the HeuristicCost plot. To someone unfamiliar with the project it might be unclear why in (a) the robot cannot proceed to the goal unless they read the caption closely. Having the hard margins as part of the image might make it more obvious.  }
% %   }
%   \label{fig:expert_traj}
%   \vspace{-1em}
% \end{wrapfigure}
%%%%%%%%%%%%%%%%%%%%%%%%%%%%%%%%%%%%%%%%%%%%%%%%%%%%%%%%%%%%%%%%%%%%%%%%%%%%%%%%%%%%%%%%%%%%%%%%%%%

We select $\bar v = 0.6$ m/s, $\bar \omega = 1.1$ rad/s, and $\bar a = 0.4$ m/s$^{2}$ to match the specifications of the Turtlebot 2 used in the hardware experiments (Sec. \ref{sec:hardware}). %Thus for the state and control constraints, we .
% For disturbances in the dynamics, 
% We set $\bar d_{xy} = 0.05$ m/s as the tracking error for the robot, and $\bar d_{\phi} = 0.15$ rad/s to account for control noise.
We set $\bar d_{xy} = 0.05$ m/s and $\bar d_{\phi} = 0.15$ rad/s to account for prediction errors, and $\alpha = 30$ to prioritize collision avoidance.
All expert trajectories are generated according to Section \ref{sec:waypt_generation}, where the replanning is done every $1.5$s to collect training data.
In Fig. \ref{fig:traj_plots} (a) to (c), we compare the expert trajectories obtained by \oldcost and \ourcost. 
% The reachability expert is able to choose better combinations of $(x,y,\theta, v)$ as waypoints. 
The reachability expert uses the full system dynamics for optimizing waypoints. As a result, it maintains an appropriate orientation and speed when going through the narrow openings.
Moreover, due to the presence of disturbances, it takes a conservative path, always staying near the middle of narrow openings, resulting in collision-free trajetcories even when there is prediction error.
In contrast, \oldcost takes a greedier path to approach the goal. To  address CNN prediction error, \cite{bansal2019combining} use an obstacle padding which makes narrow openings impossible to enter. 
%In this situation, reachability expert demonstrates its superior advantages in the trajectories.

\subsection{Test Results}
% 
%%%%%%%%%%%%%%%%%%%%%%%%%%%%%%%%%%%%%%%%%%%%%%%%%%%%%%%%%%%%%%%%%%%%%%%%%%%%%%%%
% A table present the statistics for different neural networks with control horizon = 0.25
\begin{table}
\caption{\textbf{Quantitative Comparisons in Simulation:} 
% To be written
We compute four metrics-success rate, average time to reach the goal, acceleration and jerk-on 200 navigation tasks with a replan frequency of 4Hz. 
The proposed method, WayPtNav-\ourcost,  is most successful at completing novel navigation tasks. 
% \oldcost takes the shortest time to reach the goal, but only succeeds on 11 percents fewer of the tasks.
Without the disturbances incorporated in the dynamics, \ourcost takes the shortest time to reach the goal, but the success rate largely drops because of prediction errors. 
For E2E learning, the success rate is generally lower and the trajectories are less smooth (indicated by high average jerk).}
\label{table:statistics_metrics}
\centering
\resizebox{1.0\linewidth}{!}{
\begin{tabular}{lcccc}
\toprule
\textbf{Agent} & %\textbf{Input}%
\textbf{Success (\%)} & \textbf{ Time taken (s)} & \textbf{Acceleration (m/s$^{2}$)} & \textbf{Jerk (m/s$^{3}$)} \\ \midrule
% Expert & Full map &100 &      10.78 \textpm 2.64 &    0.11 \textpm 0.03 &    0.36 \textpm 0.14 \\
% \midrule
WayPtNav-\ourcost & \textbf{63.82} &    21.00 \textpm 8.00 &    \textbf{0.06 \textpm 0.01} &    \textbf{0.94 \textpm 0.13}\\
WayPtNav-\oldcost & 52.26 &    18.82 \textpm 5.66 &    0.07 \textpm 0.02 &    1.06 \textpm 0.15 \\
WayPtNav-\ourcostNoDist & 49.24  &    \textbf{16.19 \textpm 4.8} &    0.07 \textpm 0.01  &     0.98 \textpm 0.16  \\
\midrule
E2E-\ourcost  & 8.04 &  19.55 \textpm 4.72 &   0.07  \textpm 0.01 &   2.16  \textpm 0.30 \\
E2E-\oldcost & 31.66 &   25.56  \textpm 9.85 &   0.26  \textpm 0.06 &    9.06 \textpm 1.94 \\
% \midrule
% Mapping  & Depth + Spatial Memory & 97.85 &  10.95 \textpm 2.75 &    0.11 \textpm 0.03 &    0.36 \textpm 0.14 \\
\bottomrule
\end{tabular}}
\vspace{-1em}
\end{table}
%%%%%%%%%%%%%%%%%%%%%%%%%%%%%%%%%%%%%%%%%%%%%%%%%%%%%%%%%%%%%%%%%%%%%%%%%%%%%%%%

\begin{wrapfigure}{r}{0.35\textwidth}
\vspace{-5em}
\begin{center}
\includegraphics[width=1\linewidth]{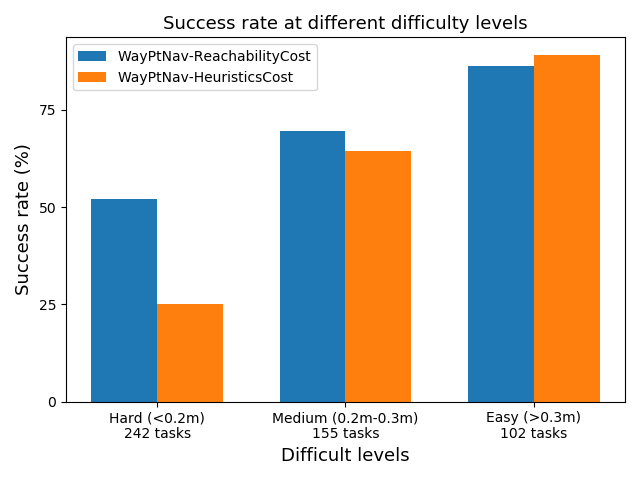}
    \end{center}
    \vspace{-2.5em}
    \captionof{figure}{Success rate in tasks of different difficulties.
    %Blue: \oldcost; Red: \ourcost; both using WayPtNav.
    Difficulties are assessed according to minimum distance to obstacles $\expertminobs$ along trajectories, divided into Hard ($\expertminobs<0.2$ m, 242 tasks), Medium ($0.2$ m $\leq \expertminobs\leq0.3$ m, 155 tasks) and Easy ($\expertminobs>0.3$ m, 102 tasks). \ourcost has significant advantage in Hard tasks.}
    \label{fig:successrate_obsdist}
    \vspace{-1.5em}
\end{wrapfigure}
We compare the different methods in Table \ref{table:statistics_metrics}. 
WayPtNav-\ourcost achieves the highest success rate and least acceleration and jerk. 
% Although \oldcost with WayPtNav takes shorter average time to reach the goal, it does not succeed in more difficult tasks that requires a longer time to complete.
WayPtNav-\ourcostNoDist takes the shortest time to reach the goal, but experiences a notable drop on the success rate.
% \vtnote{I had to read the previous sentence 3-4 times to make sense of it. I thought you were talking about "WayPtNav-ReachabilityCost" not "WayPtNav-ReachabilityCost without disturbances". The words "without disturbances" didn't stand out since they were lowercase. Maybe we could call this baseline WayPtNav-ReachabilityCost-Without-Disturbances (capital letters) to make sure readers realize that this is an entirely separate baseline from WayPtNav-ReachabilityCost. Also clearing up the explanation in the baselines section would probably help with this as well.}
% 

E2E learning has lower success rate and more jerky trajectories in genreral. Notably, \ourcost has a significant lower success rate with E2E learning compared to WayPtNav.

% Simulation analysis
\section{Analysis of Simulation Results} \label{sec:simulation_analysis}
\begin{wrapfigure}{r}{0.35\textwidth}
\vspace{-2em}
\begin{center}
\includegraphics[width=1\linewidth]{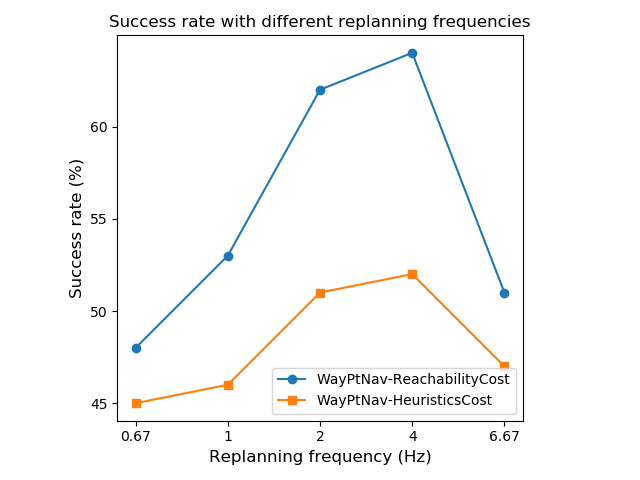}
    \end{center}
    \vspace{-2em}
    \caption{Success rates improve for both methods as the replanning frequency is increased (up to 4 Hz). \ourcost benefits considerably more with higher replan frequencies.}
    \label{fig:successrate_replan_freq}
    \vspace{-1em}
\end{wrapfigure}
% 
% 
% \subsection{Comparison with WayPtNav-\oldcost} \label{sec:reachability_cost}
\textbf{Comparison with WayPtNav-\oldcost.} Shown in Sec.~\ref{sec:expert_performance}, the reachability expert is less greedy and more robust, enabling it to navigate through cluttered environments and narrow openings. 
This results in a higher success rate during test time, as shown in Fig. \ref{fig:successrate_obsdist}, where we compare the two methods on 500 navigation tasks with varying difficulties. 
Here, we measure the difficulty of a task by the opening size that the robot must navigate through on its way to the goal.
\ourcost has much higher success rate on the ``Hard'' level, indicating that it makes robot more adept at maneuvering in narrow environments.
\smallskip
\\
% % Simulation results
% \subsection{Effect of Replanning Frequency}
\textbf{Effect of Replanning Frequency.} Replanning frequency represents how often the robot predicts new waypoints.
% based on the current image input.
Fig.~\ref{fig:successrate_replan_freq} shows the success rate on 200 navigation tasks for 5 different replanning frequencies: 0.67 Hz, 1 Hz, 2 Hz, 4 Hz and 6.67 Hz.
As the frequency increases, the success rate elevates to the peak at 4 Hz for both experts. 
Above 4 Hz, the success rate drops dramatically.

In general, a higher replanning frequency helps the system react faster to the unknown environment; however, if it is too high, the lack of visual memory in the WayPtNav framework results in myopic decisions, leading to a drop in the success rate.
\ourcost benefits more from the higher replanning frequencies compared to \oldcost, as the greedy behaviors from the latter tend to drive the robot to cut corners, often leading to situations that are hard to recover from.
\smallskip
\\
% \subsection{Effect of Adding Disturbances} \label{sec:adding_disturbances}
\textbf{Effect of Adding Disturbances.} Adding disturbances in the dynamics is crucial for improving test performance: it not only accounts for dynamics uncertainties, but also models neural network prediction errors. 
Without disturbances, trajectory can be ``too optimal'' so that a little deviation of robot's state or minor errors from the CNN results in collision. 
This is evident from Table \ref{table:statistics_metrics}, where the success rate drops from $63.82\%$ to $49.24\%$ in the absence of disturbances.

\begin{figure}
    % \vspace{5pt}
    \centering
    \includegraphics[width=0.89\linewidth]{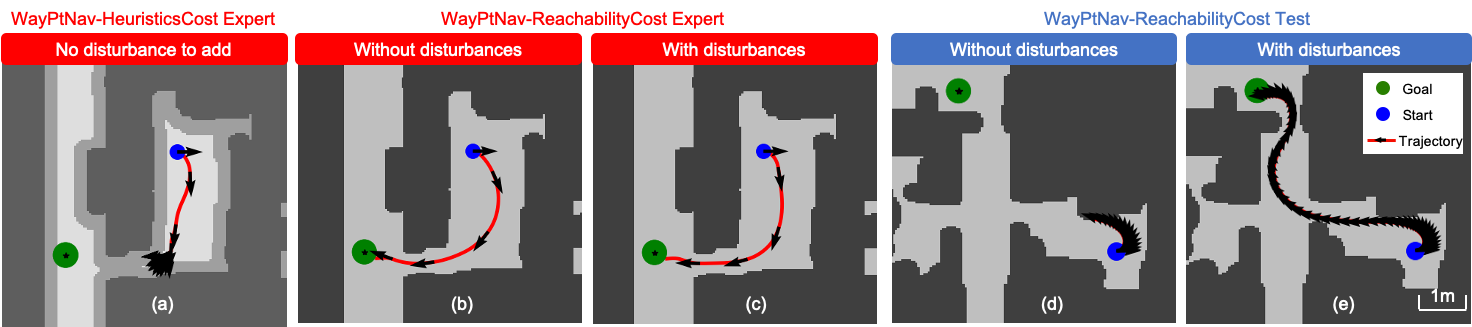}
    % \caption{Comparison of trajectories with/without disturbances in dynamics. (a) and (b): Expert trajectories; (c) and (d): Test trajectories. With disturbances incorporated in the dynamics, the reachability expert is more likely to stay in the middle of the road. This ability is transferred to test times, where the robot with disturbances added is more resistant to the prediction errors from the neural network and manages to take a collision free path to the goal.}
    \vspace{-0.5em}
    \caption{Trajectory comparison. (a), (b), and (c) are expert trajectories; (d) and (e) are test trajectories. In (a), the baseline expert starts with a greedier path and fails to enter the narrow opening due to hard obstacle padding (medium grey) used during waypoint optimization. Comparing (b) and (c), the reachability expert can safely go through the openings, but is more likely to stay in the middle of the road with disturbances incorporated in dynamics.
    This ability is transferred to test times in (d) and (e), where the robot with disturbances added is more resistant to the prediction errors from the neural network and manages to take a collision free path to the goal.}
    \label{fig:traj_plots}
\vspace{-1em}
\end{figure}

We examine reachability expert trajectories with and without the disturbances in Fig. \ref{fig:traj_plots} (b) and (c). 
Without disturbances, the expert chooses a path close to the obstacles, while with disturbances, the expert stays near the middle of the road.
Although both experts succeed in reaching the goal, disturbances lead to more robust trajectories, which translate to test time as shown in Fig. \ref{fig:traj_plots} (d) and (e)).
Without disturbances, the robot tries to avoid the wall but fails, while with disturbances, the robot is able to stay in the middle of the opening, and pass through a very narrow doorway.
\smallskip
\\
% \subsection{Comparison with E2E Learning} \label{sec:e2e_comparison}
\textbf{Comparison with E2E Learning.} Our conclusions here are consistent with the findings in \cite{bansal2019combining} -- the model-based approach (WayPtNav) leads to a higher success rate and significantly smoother trajectories.
The success rate of E2E learning declines further for the reachability expert as the control profiles are even more nuanced, making it challenging to learn them.
\smallskip
\\
% \subsection{Failure Cases} \label{sec:failure_case}
\textbf{Failure Cases.} Since we do not construct a map or have any visual memory in this framework, the robot struggles in the navigation scenarios where it needs to ``backtrack''. 
In addition, when the room layout is too different from the training time, the CNN fails to predict good waypoints.
% %%%%%%%%%%%%%%%%%%%%%%%%%%%%%%%%%%%%%%%%%%%%%%%%%%%%%%%%%%%%%%%%%%%%%%%%%%%%%%%%%%%%%%%%%%%%%%%%%%%
\section{Hardware Experiments} \label{sec:hardware}
% \begin{figure}
%     \vspace{5pt}
%     \centering
%     \includegraphics[height = 0.3\linewidth]{figures/Scenarios_2_3_Big_Hotdog.PNG}
%     \vspace{-1em}
%     \caption{Experimental scenarios 2 and 3. Scenario 2 (top), and 3 (bottom) are shown in 3rd and 1st person along with a birds-eye view. The goal position is marked by a green ellipsoid, and a the trajectory taken by the \ourcost CNN that goal is shown in red. The Turtlebot's starting location is shown as a blue circle.}
%     \label{fig:Scenarios_2_3}
% \vspace{-1em}
% \end{figure}

We tested our framework on a Turtlebot 2 hardware testbed (Fig. \ref{fig:Scenario_1}), using monocular RGB images from the onboard camera for navigation.
% 
% Hardware experiments were executed using a Turtlebot 2 testbed. RGB input images were obtained using an Intel\textregistered\ Realsense\texttrademark\ D435 camera (69.4\textdegree\ x 42.5\textdegree\ x 77\textdegree\ RGB FOV) mounted 1 meter above and pointed 30\textdegree\ below the horizontal plane. Neural network and planning computations were executed using an on-board computer (Intel\textregistered\ i7-9750H, RTX 2070 w/ Max-Q Design). The computer interfaced the Turtlebot and camera via two USB-C cables. \newline
% 
%%%%%%%%%%%%%%%%%%%%%%%%%%%%%%%%%%%%%%%%%%%%%%%%%%%%%%%%%%%%%%%%%%%%%%%%%%%%%%%%%%%%%%
% Scenario 2/3 Fig.        
%%%%%%%%%%%%%%%%%%%%%%%%%%%%%%%%%%%%%%%%%%%%%%%%%%%%%%%%%%%%%%%%%%%%%%%%%%%%%%%%%%%%%%
% 
%%%%%%%%%%%%%%%%%%%%%%%%%%%%%%%%%%%%%%%%%%%%%%%%%%%%%%%%%%%%%%%%%%%%%%%%%%%%%%%%%%%%%%
% 
We tested the \ourcost and \oldcost CNNs directly on the Turtlebot without any additional training or fine-tuning.
% Depth images taken by the D435 camera any CNN.

\begin{wrapfigure}{r}{0.35\textwidth}
\vspace{-2em}
    \begin{center}
        \includegraphics[width=\linewidth] {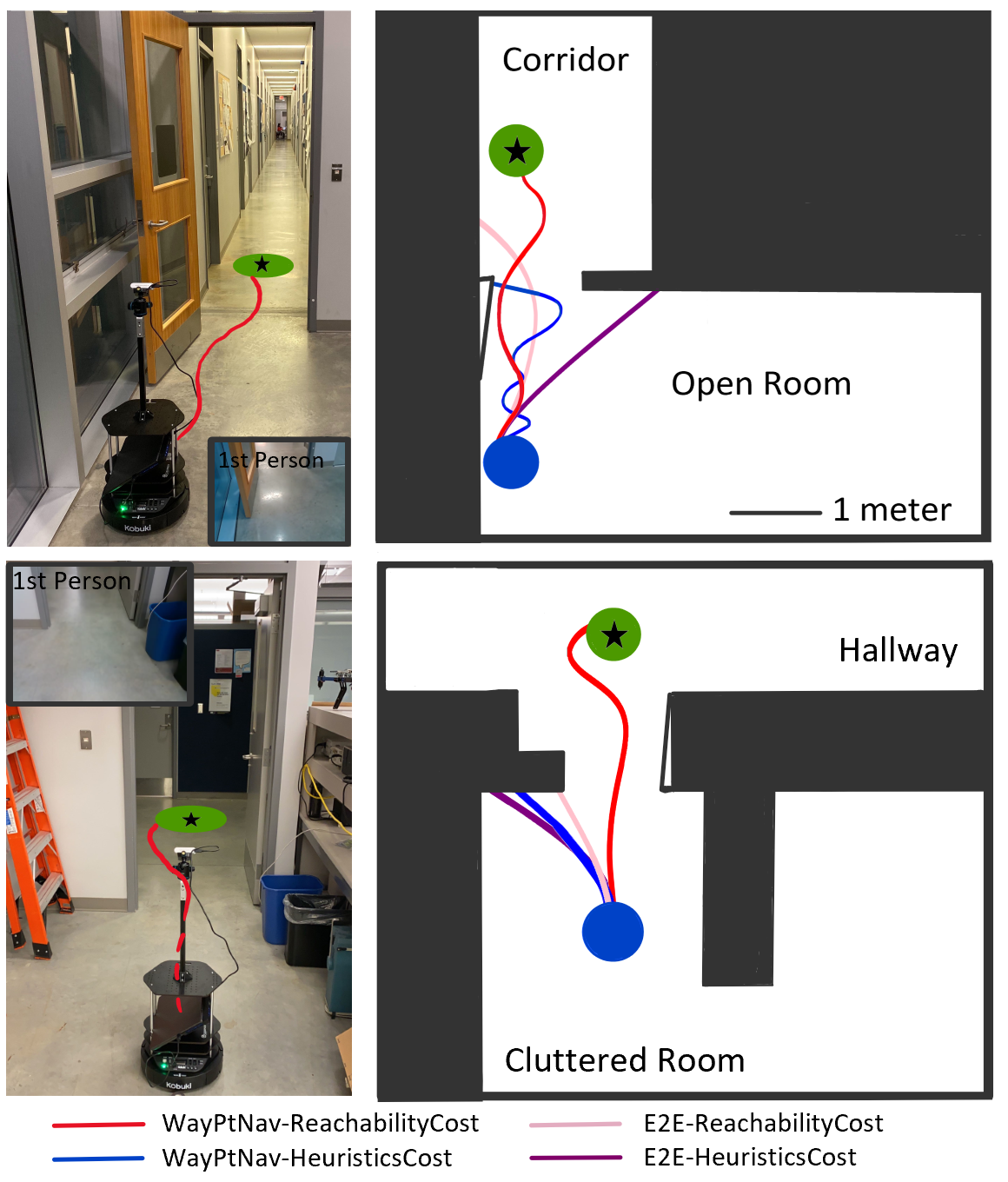}
    \end{center}
    \vspace{-1em}
    \caption{Experimental scenarios 2 (top) and 3 (bottom), shown in 3rd and 1st person along with birds-eye view. The goal is marked in green, and the trajectory taken by the \ourcost CNN to the goal is shown in red. The Turtlebot's starting location is shown as a blue circle.}
    \label{fig:Scenarios_2_3}
    \vspace{-2.5em}
\end{wrapfigure}

Experiments were carried out in 3 separate areas of Simon Fraser University, each of which were absent from the training set. An outline of each experiment and the trajectories taken by the Turtlebot with each CNN are shown in Figures \ref{fig:Scenario_1} and \ref{fig:Scenarios_2_3}. Video footage of all experiments can be found at \url{https://www.youtube.com/playlist?list=PLUBop1d3Zm2uDGGfGrjWiSjrSlzo5vWMs}.

Each scenario required the Turtlebot to traverse narrow spaces and doorways, both of which have shown a low success rate for the \oldcost CNN. For the first scenario, the Turtlebot maneuvered through a narrow doorway, and then around the nook behind the door to reach the goal (Fig. \ref{fig:Scenario_1}). The second and third scenarios required the Turtlebot to move from an open room into a narrow corridor and from a cluttered environment into a hallway respectively (Fig. \ref{fig:Scenarios_2_3}). 

For each scenario, the \oldcost CNN was unable to maneuver through the doorways, and collided with the wall at full speed, while the \ourcost CNN successfully navigated through the doors and reached the goal. The ability to navigate through narrow environments and doorways is a key improvement seen in the \ourcost CNN. For both methods, the neural networks trained using end-to-end learning were unable to reach the goal. 

% Discussions on disturbances
\section{Discussion}

The CNN prediction error is the deviation of the predicted waypoint away from the optimal one, resulting in a deviation of the robot trajectory. 
In the TTR and TTC value computations, worst-case disturbances are added to the dynamics, which lead to conservative TTR values (larger) and TTC values (smaller) that account for possible deviations in the robot trajectory.
This conservatism can effectively compensate for the prediction error.

Currently we choose the disturbance bound based on empirical results from our experiments.
In the future, more accurate bound can be learned using principled analytical and data-driven methods.

% Conclusion
\section{Conclusion}

In this paper, we present a novel method to generate training data for waypoint prediction in visual navigation, using reachability analysis.
Our method helps the neural network to learn to predict efficient and safe waypoints given system dynamics and observations of the environment.
We also use disturbances in dynamics to model neural network prediction errors and greatly enhance its robustness.
Simulation and real robot experiments demonstrate higher success rate and smoother trajectories in navigation tasks with our method, which crucially enables the robot to pass through narrow passages such as doorways.
Immediate future work includes adding memory in our navigation framework, investigating the data mismatch between training and test scenarios to better transfer the expert performance in test time, and incorporating human agents.

% Acknowledgments---Will not appear in anonymized version
\acks{This work was funded in part by the Natural Sciences and Engineering Research Council of Canada (NSERC) Discovery Grants Program, by the DARPA Assured Autonomy program, and by the SRC CONIX program.}

\bibliography{bibliography.bib}

\end{document}